\documentclass[journal]{IEEEtran}
\usepackage[utf8]{inputenc}
\usepackage{soul}
\usepackage{lettrine}
\usepackage{todonotes}
\usepackage{cite}
\usepackage{supertabular,booktabs}
\usepackage{float}
\usepackage{comment}
\usepackage{makecell}
\usepackage{graphicx}
\usepackage{pdfpages}
\usepackage{amsmath}
\usepackage{algorithm}
\usepackage{algpseudocode}
\usepackage{subfig}
\usepackage{amssymb}
\usepackage{relsize}
\usepackage{subfiles}

\newcolumntype{L}{>{\centering\arraybackslash}m{6.5cm}}

\makeatletter
\newcommand\fs@betterruled{%
  \def\@fs@cfont{\bfseries}\let\@fs@capt\floatc@ruled
  \def\@fs@pre{\vspace*{8pt}\hrule height.8pt depth0pt \kern2pt}%
  \def\@fs@post{\kern2pt\hrule\relax}%
  \def\@fs@mid{\kern2pt\hrule\kern2pt}%
  \let\@fs@iftopcapt\iftrue}
\floatstyle{betterruled}
\restylefloat{algorithm}

\usepackage{hyperref}
\hypersetup{hidelinks}
\usepackage{orcidlink}

\def\BibTeX{{\rm B\kern-.05em{\sc i\kern-.025em b}\kern-.08em
    T\kern-.1667em\lower.7ex\hbox{E}\kern-.125emX}}
    
\algrenewcommand\algorithmicrequire{\textbf{Input:}}
\algrenewcommand\algorithmicensure{\textbf{Output:}}
\algnewcommand\algorithmicforeach{\textbf{for each}}
\algdef{S}[FOR]{ForEach}[1]{\algorithmicforeach\ #1\ \algorithmicdo}

\title{
Warmup and Transfer Knowledge-Based Federated Learning Approach for IoT Continuous Authentication}

\author{
    \IEEEauthorblockN{Mohamad Wazzeh$^1$, Hakima Ould-Slimane$^2$, Chamseddine Talhi$^1$, Azzam Mourad$^{3,4\orcidlink{0000-0001-9434-5322}}$ and Mohsen Guizani$^{5\orcidlink{0000-0002-8972-8094}}$}\\
    
    \IEEEauthorblockA{$^1$Department of Software and IT engineering, École de Technologie Supérieure (ÉTS), Canada}\\
    \IEEEauthorblockA{$^2$Department of mathematics and computer science, Universite de Quebec a Trois-Rivieres (UQTR), Canada}\\
    \IEEEauthorblockA{\normalsize$^3$Cyber Security Systems and Applied AI Research Center, Department of CSM, Lebanese American University, Lebanon}\\
    \IEEEauthorblockA{$^4$Division of Science, New York University, Abu Dhabi, UAE}\\
    \IEEEauthorblockA{$^5$Mohammad Bin Zayed University of Artificial Intelligence, Abu Dhabi, UAE}\\
    \IEEEauthorblockA{
    \href{mailto:mohamad.wazzeh.1@ens.etsmtl.ca}{mohamad.wazzeh.1@ens.etsmtl.ca},
    \href{mailto:ould.hakima@gmail.com}{ould.hakima@gmail.com},
    \href{mailto:chamseddine.talhi@etsmtl.ca}{chamseddine.talhi@etsmtl.ca},
    \href{mailto:azzam.mourad@lau.edu.lb}{azzam.mourad@lau.edu.lb},
    \href{mailto:mohsen.guizani@mbzuai.ac.ae}{mohsen.guizan@mbzuai.ac.ae}
    }
}

\begin{document}
\maketitle

\begin{abstract}
Continuous behavioural authentication methods add a unique layer of security by allowing individuals to verify their unique identity when accessing a device. Maintaining session authenticity is now feasible by monitoring users' behaviour while interacting with a mobile or Internet of Things (IoT) device, making credential theft and session hijacking ineffective. Such a technique is made possible by integrating the power of artificial intelligence and Machine Learning (ML). Most of the literature focuses on training machine learning for the user by transmitting their data to an external server, subject to private user data exposure to threats. In this paper, we propose a novel Federated Learning (FL) approach that protects the anonymity of user data and maintains the security of his data. We present a warmup approach that provides a significant accuracy increase. In addition, we leverage the transfer learning technique based on feature extraction to boost the models' performance. Our extensive experiments based on four datasets: MNIST, FEMNIST, CIFAR-10 and UMDAA-02-FD, show a significant increase in user authentication accuracy while maintaining user privacy and data security.
\end{abstract}


\section{Introduction}
Maintaining the authenticity of a device is crucial in guaranteeing ownership and providing access to a legitimate owner \cite{dbouk2019novel}. Current authentication strategies are doable by matching a pattern, entering a password, using two-factor authentication, and others \cite{mourad2015sba, jebbaoui2015semantics}. These techniques provide an authentication mechanism between users and their devices. However, traditional methods failed to address the continuity of session authenticity. For example, suppose a user logged in to a device using his traditional authentication technique and left the session unattended. A malicious user or system can take over the session and the device. In addition, any credential exposure may put the device at risk of accessing private user data \cite{hammoud2018detection}. One way to resolve this limitation is to ask users to input credentials frequently. However, it would interrupt the user workflow, which is not practical. 

Continuous authentication is an advanced security system that can implicitly and continuously authenticate user devices in the background by collecting sensor behavioural data from the user's device and processing these data with a machine-learning algorithm. Such a procedure is achievable by collecting integrated sensors available on their devices, such as smartphones or IoT sensors and transferring them to an external server where a machine learning algorithm trains these data to create an authentication model for each device. The sensors' data collection and processing of IoT devices \cite{chehab2020lp} is done in the background, and the verification is doable while the user interacts with a device. A machine learning algorithm uses these data to evaluate whether the user is an imposter who gains access to the system or a legitimate owner trying to complete a task using his device.   

\begin{figure}[ht]
  \centering
\includegraphics[scale=.30]{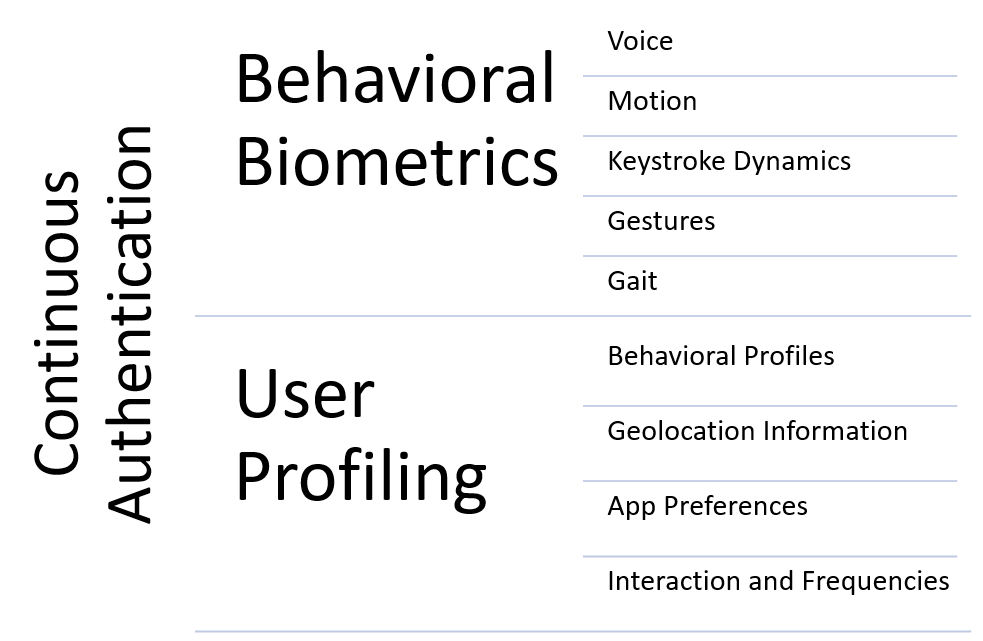}
  \caption{Behavioural biometric-based authentication modalities.}
  \label{fig:CA_modalities}
\end{figure}

IoT and smartphone devices \cite{salahuddin2018softwarization, mourad2020ad} have built-in hardware  (e.g. camera, microphone, accelerometer, gyroscope, etc.) and software (e.g. Global Positioning System (GPS), applications statistics, Proximity, Geomagnetic, etc.) based sensors. Moreover, Figure \ref{fig:CA_modalities} shows the different modalities of behavioural biometrics and user profiling. The \textit{Behavioral Biometrics} refers to the usage of device-integrated sensors, while \textit{User Profile} refers to the devices' applications and the user preferences in those applications. The conventional continuous authentication approaches such as the works in \cite{samet2019touchmetric, lee2017implicit, 4_zhu2019riskcog} used mobile device internal sensors to capture personal user data. Afterwards, the captured data are transmitted into an external server, where a machine learning algorithm creates a model that could identify user behaviour based on his/her usage of the device. However, collecting raw personal data raises concerns regarding data transmission security and storage safety. In addition, the General Data Protection Regulation (GDPR) \cite{2_yang2019federated} imposed by the European Union has introduced strict regulations on data sharing and techniques to protect user privacy, which has created new challenges for data collection. Some strategies exist in \cite{5_valero2020machine} and \cite{6_jorquera2018improving}, which consider the training of user data on the device without communication with an external server. However, their approaches did not lead to high authentication results and showed high computational and energy costs for mobile devices. Their practices use anomaly detection algorithms to train an ML model for each user, where negative data samples are not needed to create the models. 

This work aims to deliver a framework that can protect the user's privacy and allow them to continuously authenticate their identity with mobile and IoT devices. Behavioural authentication is possible by leveraging built-in sensors to continuously inspect the behavioural usage of users' devices in the background. In the literature, some work targets preserving private user data. Among them, the authors in Google \cite{8_mcmahan2017communication} suggest Federated Learning, a distributed machine learning strategy that can learn from several users while keeping their data private. In such a technique, the model weights are used to replace the need for users' raw data to assemble a new ML model to predict the data labels. In their approach, the user builds a local model trained from the collected data and then communicates the model weights to an external server. The latter collects and aggregates models from each participating user to obtain a robust model containing all users' knowledge. The authors proposed a federated averaging algorithm (FedAvg) that averages the weights of the users' models. However, such a FL technique imposes problems for the continuous authentication system that we must address. 
Continuous authentication collected data such as gait, motion, voice, and others have a unique behavioural fingerprint for every user. The data collected by users represent their distinctive behavioural identity while interacting with each device. Therefore, the data are linked to a unique class or label to process the machine-learning model for each user. In addition, the data collected vary in terms of data, as users' data collection duration, device frequency and activities differ from user to user. Hence the behavioural data collected from the devices' sensors are not independent nor identically distributed (non-IID). The usage of such data has a drop in the accuracy of the models when used in a federated learning architecture \cite{wahab2021federated, abdulrahman2020survey, abdulrahman2020fedmccs}. This deterioration is caused by the different models' weight of the clients and the aggregated model. Since in behavioural data, every user collects data for his use. Therefore each user has access to only one label for their collected data.

Most of the literature on continuous authentication has not discussed the possibility of federated learning architecture for continuous authentication. Only a handful of works considered a FL possibility in the problem we are addressing. For example, the work in  \cite{9_oza2021federated} used a mix of split learning and federated learning technique to handle the non-IID problem of continuous authentication. However, the authors still share users' mean and variance data with the server. As well as in another work in \cite{monschein2021towards}, the authors transmit raw data to each server independently. Alternatively, the works of \cite{wazzeh2022privacy, 14_zhao2018federated} used a warmup approach by sharing a small amount of data with a server to create a booster model for the federated learning model. Hence, motivated by the discussed works, we present a novel federated learning technique that employs transfer learning as a booster model to alleviate the problem of Non-IID and weight divergence presented in continuous authentication. The transfer learning technique in this work will be used as a feature extractor for the model to increase performance and reach convergence with fewer rounds.
Furthermore, we implement a centralized training approach to test our proposed framework, where all user behaviour data resides on one server, and we compare the FL results. Our contributions are recapped as follows:
\begin{itemize} 
\item Build a novel strategy for continuous authentication based on federated learning which enables the preservation of private user data.
\item Elaborate on a transfer learning mechanism as a feature extractor for continuous authentication models.
\item Perform extensive experiments on various well-known datasets, including continuous authentication datasets, to further evaluate our approaches' performance.

\end{itemize}

In section II, we present the related work that discussed the usage of Federated Learning and Transfer Learning with continuous authentication. Next, section III presents our framework and explains the transfer learning step with the warmup technique. In section IV, we show the experimentations on each dataset we used to evaluate the performance of our framework and provide an analytical review of them. Finally, in section V we conclude the work presented in this paper and suggest future work that could improve the proposed framework.

\section{Related Work}
\begin{table*}[ht]
\caption[Comparison of Continuous Authentication approaches in the literature.]{Comparison of Continuous Authentication approaches in the literature. \newline 
    Ac: Accelerometer. Gy: Gyroscope. Gr: Gravity. Ma: Magnetometer. Ts: Touch Screen. Lo: Location. MP: Multilayer Perceptron. IF: Isolation Forest. SVM: Support Vector Machine. KRR: Kernel Ridge Regression. LSTM: Long Short-Term Memory. FAR: False Acceptance Rate. TAR: True Acceptance Rate. FRR: False Rejection Rate. EER: Equal Error Rate.}
\label{tab:table_ML_Summary}
\centering
\renewcommand{\arraystretch}{2}
\resizebox{\textwidth}{!}{%
\begin{tabular}{|c|c|c|c|c|c|}
\hline
\textbf{Publication} &
  \textbf{Features Level} &
  \textbf{ML} &
  \textbf{Participants} &
  \textbf{Accuracy} &
  \textbf{Authentication time} \\ \hline
\cite{samet2019touchmetric}   - 2019 &
  Ts &
  MP &
  34 &
  100 &
  NA \\ \hline
\cite{5_valero2020machine} - 2020 &
  App   Usage, Lo &
  IF &
  NA &
  89 &
  Real-time \\ \hline
\cite{4_zhu2019riskcog} - 2020 &
  Ac, Gy, Gr &
  SVM &
  1,513 &
  \begin{tabular}[c]{@{}c@{}}Accuracy:   95.6, \\      TAR: 73.28\end{tabular} &
  3.2s \\ \hline
\cite{sun2020kollector} - 2020 &
  Ac,Ts &
  MVB &
  26 &
  \begin{tabular}[c]{@{}c@{}}Accuracy:   94.24, \\      EER: 8.42\end{tabular} &
  0.001s \\ \hline
\cite{6_jorquera2018improving} - 2018 &
  Ac,Gy,App   Usage &
  IF &
  50 &
  82.5 &
  3-5s \\ \hline
\cite{lee2017implicit} - 2017 &
  Ac, Gy &
  KRR &
  35 &
  \begin{tabular}[c]{@{}c@{}}Accuracy:   98.1, \\      FAR: 2.8, \\      FRR: 0.9\end{tabular} &
  6s \\ \hline
\cite{gascon2014continuous} - 2014 &
  Ts &
  SVM &
  300 &
  FAR:   1, TAR: 92 &
  NA \\ \hline
\cite{abuhamad2020autosen} - 2020 &
  Ac,Ma,Gy &
  LSTM &
  84 &
  \begin{tabular}[c]{@{}c@{}}FAR:   0.95, \\      FRR: 6.67, \\      EER: 0.41\end{tabular} &
  0.5s,   1s \\ \hline
\cite{sanchez2021authcode} - 2020 &
  \begin{tabular}[c]{@{}c@{}}Computer   (Cp and App Usage), \\      Smartphone(Ac, Gy and App Usage)\end{tabular} &
  \begin{tabular}[c]{@{}c@{}}MP,   XGBoost,\\       RF and LSTM\end{tabular} &
  5 &
  \begin{tabular}[c]{@{}c@{}}Precision:   99.32, \\      Recall: 99.33,\\      F1-Score: 99.33\end{tabular} &
  2s \\ \hline
\end{tabular}%
}
\end{table*}

This section outlines the continuous authentication related work and its perspectives in federated learning and transfer learning contexts.
\par \hspace{0pt} \\ \indent
 % \subsection{Centralized Continuous Authentication} 
\textit {Continuous authentication} can be achieved by continuously authenticating a user's device using the user profile and behavioural sensor data to create a unique profile that could verify users by linking user-devices behaviour relations. The literary work applied continuous authentication as a mechanism on mobile and IoT devices for the continuity of session security. Several traditional centralized approaches \cite{ sanchez2021authcode, lee2017implicit, 4_zhu2019riskcog, gascon2014continuous, abuhamad2020autosen}, discussed the feasibility of training machine learning algorithms on the server side by sending all raw data to an external server for further processing. The results from these works demonstrated decent authentication scores and could work on a user's smartphone without affecting its usability. However, all these strategies lack the basis for maintaining user data security and privacy since raw sensor data are shared from user devices to an external server where further data processing occurs.
\par \hspace{0pt} \\ \indent
\textit {On-Device training}. The resources of mobile devices have increased in processing, storage and storage. Therefore, it is possible to manage device data for processing and evaluation \cite{rahman2020internet}. Different approaches in \cite{5_valero2020machine, 6_jorquera2018improving} used anomaly detection techniques to train ML models on the device itself. Authors in these works used models that do not require high resource consumption to make the approach feasible on low-end devices, where the models are trained using only the positive samples of the device. Analysis of such technique shows relatively lower accuracy scores and high false positives as the models lack the negative data sample in the training process.
\par \hspace{0pt} \\ \indent
\textit {Federated Learning.} The privacy and security of the data are crucial for most individuals and organizations \cite{elayan2021sustainability}. In order to benefit from the global knowledge of individual models, FL is used to train a shared model among several nodes. The decentralization architecture allows the training of a global model between the collaborators. The user can then share with the server the model weight instead of sharing his raw private data. Shared weights from each user are combined on the server to extend the learning process between individuals and create a global model in multiple rounds until the models converge. Limited studies have been conducted on using FL in continuous behavioural authentication. For instance, a study in \cite{14_zhao2018federated} creates a warmup model to boost the initial weights by transmitting a tiny percentage of data to other participants and the server. The clients' transferred data are shared with the server and the user devices. Their technique generated an accuracy improvement of 30\% for the CIFAR10 dataset. However, such advancement comes at the expense of communicating data between the users' devices.

The work of \cite{9_oza2021federated} used an approach that incorporates federated and split learning in a single round of communications. They employed a trained model's pre-existing weights and sent statistical user data with a remote server. In a separate work in \cite{monschein2021towards}, a federated approach is considered for peer-to-peer learning to address the problem of continuous authentication. The authors create federated learning between servers instead of a FL between the clients. However, their approach shares the clients' raw data with the servers. In recent work in \cite{wazzeh2022privacy}, a federated learning architecture for continuous authentication is modelled on demo datasets using the FedAvg algorithm. Using a small data cut from users, the authors employed a strategy to stimulate the initial model weights, thus increasing the model's performance. However, the datasets tested on the framework were not authentication datasets, and the sparsity of the data was not addressed in their work.
In summary, it is possible to apply federated learning in continuous authentication. It has promising potential to keep user data private and offload large loads from centralized servers. Minimal works discuss the potential of using federated learning with such a unique case of continuous authentication. To our knowledge, no other work has addressed this feature for authenticating user devices using their behavioural data. 
\par \hspace{0pt} \\ \indent
\textit {Transfer Learning.}
Machine learning algorithms often need enormous data samples to perform well. The interaction of individuals with IoT devices captures biometric behaviour data from these devices' sensors. These different modalities and interactions of the devices are presented in Figure \ref{fig:CA_modalities}. Although each user has a unique way of interacting, the interaction type and device sensors remain similar in most cases. Transfer learning is an approach where the knowledge generated by a model can be transferred when dealing with similar problems. It can be used when the training data available are insufficient to have a ML model that can attain a high-performance score. In addition, transferring knowledge of a particular model developed for a task can be used initially for a model on another task \cite{weiss2016survey}, which will lessen the resources cost of training and increase the robustness of the ML model. In another work done in \cite{kong2020continuous}, the authors perform a continuous authentication technique by leveraging the gesture interaction with IoT devices in a smart home environment. They used a transfer learning technique for a cross-domain adaptation to avoid training the model in various environments, reducing the resource consumption for the task at hand. They showed that using such a knowledge basis model, they could maintain a high accuracy performance with fewer data samples. In another work \cite{he2022gait2vec}, the authors created a continuous authentication approach for smartphone users using their sensor data for the gait modality. They used transfer learning as a feature extractor to facilitate the training duration and enhance the robustness of the ML model. The trained model extracted distinguishable features and provided better accuracy scores when tested in different usage environments. 

We have summarized the work of the literature in table \ref{tab:table_ML_Summary} where we show each of the works with their machine learning aspects, the number of participants that were collecting data and the metrics the authors used to evaluate their work performance.

\begin{figure*}[htp]
  \centering
  \includegraphics[width=.95\textwidth]{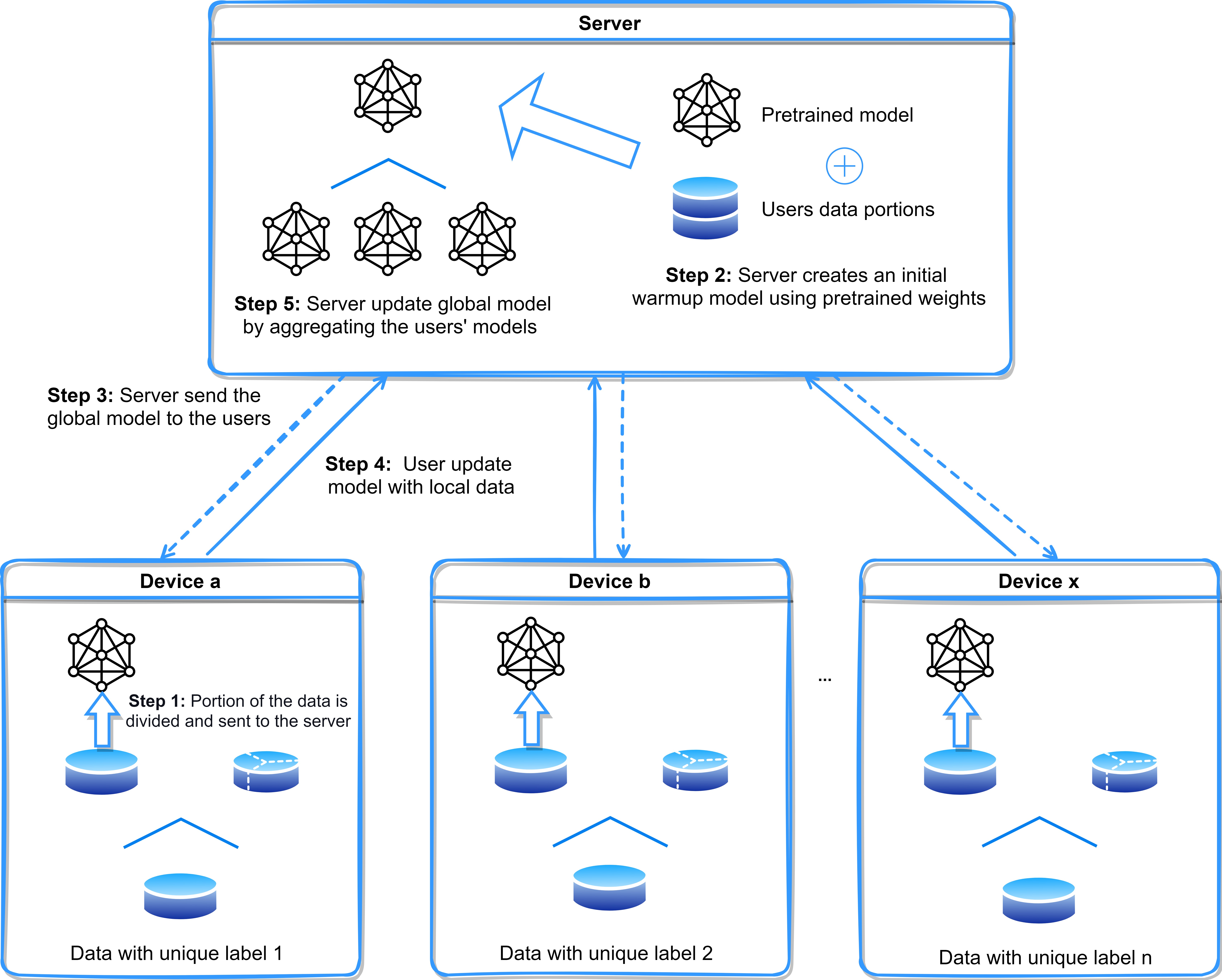}\hfill
    \caption{Warmup and Transfer Knowledge-Based Federated Learning architecture for continuous authentication}
    \label{fig:d3_FL_tl}
\end{figure*}

\section{Proposed Framework}
This section illustrates our approach to building a continuous authentication framework based on federated learning.  

\subsection{Problem Formulation}
Each client $k$ has its dataset $D_k$ and a unique label $y_k$, where $d^k_s$ is a data sample of $D_k$ of size $s$, and $K$ is the total number of clients. The representations are as follows:

\begin{equation} \label{} 
D_k = (d^k_1, d^k_2, d^k_3, ..., d^k_s)
\end{equation} 

\begin{equation} \label{} 
d^k_s = (f^s_1 + f^s_2 + ... + f^s_F)
\end{equation} 

Where $F$ is the set of features in each $d^k_s$, every feature represents a numerical pattern part of a data sample. The input given to the neural network model is represented in a sequence of $D_k$ and the accuracy $A$ output predicted of the model accuracy for a particular client $k$, and $a_k$ is the local accuracy prediction for samples of a particular client. The representation is as follows:

\begin{equation} \label{} 
A = (a_1, a_2, ..., a_{|k|})
\end{equation} 

Where $Y$ represents the true label of the clients, and each client has only one unique label as follows:
\begin{equation} \label{} 
Y = (y_1, y_2, ..., y_k)
\end{equation} 

We refer to our transfer learning method similar to the definition in \cite{pan2009survey}, where we have a current domain $Q_S$ and a learning task $T_S$ for clients to perform. The target domain of a different dataset is represented by $Q_T$ and the learning task $T_T$. We used the transfer learning to increase the learning of the prediction target function $f_T(\cdot)$ in $Q_T$, benefiting from the knowledge in $Q_S$ and $T_S$. In addition we note that $Q_S \neq  Q_T$, or $T_S \neq T_T$.

\subsection{Non-IID and Unique Class}

 \par \hspace{0pt} \\ \indent
The classic federated averaging algorithm developed by Google  \cite{8_mcmahan2017communication} is based on data in the form of IID. Devoting such an algorithm to the problem of continuous authentication will generate poor results in model accuracy. The inferior model performance is due to the essence of the behavioural biometric data of the user. Such data have an extreme case of non-IID where users have a distinctive number of data instances and a unique label in which non of their data labels is shared with any other user. Some literary work in \cite{9_oza2021federated, 14_zhao2018federated, 15_li2021federated} explored non-IID data usage with FedAvg, where the clients have only the positive data in their devices with no access to other clients labels. The unique problem of non-IID in continuous authentication seems unrealistic in other areas. Such a non-IID data distribution poses a significant challenge since the FedAvg algorithm did not achieve optimal model convergence due to unique labels on the client side \cite{15_li2021federated}.In a study in \cite{14_zhao2018federated}, the authors study the iidiness level of the data in a FL context. They show that the different model weights variation of the clients will cause the global ML model to be prejudiced toward some clients. The data skewness of the users drives the divergence in models' weights. The exact problem was also discussed in another study in \cite{15_li2021federated}, where the authors investigated various FL algorithms' performance with the FedAvg. They emphasized the depth of the non-IID issue and showed how it could negatively affect the global performance of the models. The authors, in their experiments, showed that the unique skewed distribution of the labels has the highest performance degradation of the models. Another work also mentioned the problem in \cite{15_li2021federated}. The authors analyzed several algorithms with FedAvg. Their experiments showed that the ML accuracy in the case of federated learning is not increasing when each user has only one label for the data collected.

\subsection{Implementation details}
\par \hspace{0pt} \\ \indent
To address the problems mentioned earlier regarding the weight divergence in applying federated learning for continuous authentication, we propose a novel approach that uses the transfer learning technique to train a warmup model for federated learning. The initially created model will be able to use a pre-trained weight and adapt to the similarity of the clients' data based on the feature extractor. The weights created by the warmup model will be used to reduce the impact of the weight divergence in the models. In the following steps, we summarize our proposed framework:

\begin{itemize}

\item 
\textbf{Step 1:} The user collects behavioural data while interacting with his device. The user-collected data are divided, and a small number of samples are sent to the server. This step is only needed during round one of user participation.

\item 
\textbf{Step 2:} Portions of users' data are aggregated to train a ML model centrally. This model will be an initial model to start the federated learning process. Pre-trained model weights are used as feature extractors to boost the model's learning process. We freeze a particular number of model layers to act as feature extractors and leave the remainder of the layers for the model to be trained. 

\item 
\textbf{Step 3:} After the initial model creation is completed, the server forwards the model weight to the clients.
 
\item 
\textbf{Step 4:} Every federated participant utilizes the new model weights and trains a new local model employing privately gathered data. The clients update the server with the new weights afterwards.
 
\item 
\textbf{Step 5:} The collected user model weights are aggregated by the server by applying a federated algorithm and updating the clients with the new weights. 

\end{itemize}

Our proposed approach managed the non-IID challenges when applying federated learning with the FedAvg algorithm, allowing continuous authentication for individuals' unique labels to benefit the participants. Therefore, it addresses the issue of model divergence by starting with better weights through an initial collaborative model. 

\begin{algorithm}
\caption{Warmup and transfer knowledge algorithm.}
\label{alg:fedavg_data_sharing}
\begin{algorithmic}[1]
\State \textbf{Server side:}

\State $\hskip1.0em$Create initial model or load pre-trained weights $w_0 $
\State $\hskip1.0em w_0 \leftarrow$ \textbf{Train}($-1, w_0, \zeta$) % zeta is the warmup data collected form clients 

\State $\hskip1.0em\textbf{for}$ $r = 1,2,... \textbf{ do} $

\State $\hskip2.0em p \leftarrow$ max($C \cdot K$, 1)  
\State $\hskip2.0em S_k \leftarrow $(random selected set of $k$ users)
\State $\hskip2.0em \textbf{for} $ each $k \in  S_k \textbf{ do}$

\State $\hskip3em w^k_{r+1} \leftarrow$ \textbf{TrainUser}$(k, w_r)$

\State $\hskip2.0emw_{r+1} \leftarrow  \sum ^k_{k=1} \frac{n_k}{n} w^k_{r+1}$

\State 
% D is the data
\State \textbf{Train}$(k, w, D): $
\State $\hskip1.0em \beta  \leftarrow$ (split $D$ into batches of size B)
\State $ \hskip1.0em \textbf{for} $ epoch $i$ from 1 to $E$ \textbf{ do}
\State $ \hskip2.0em \textbf{for} $ each mini-batch b $ \in \beta \textbf{ do}$
\State $ \hskip3.0em w \leftarrow w - \eta\bigtriangledown\ell(w;b) $
\State \hskip1.0em return $w$

\State 

\State \textbf{User side:}
\State $\hskip1.0em$ $\textbf{TrainUser}(k, w): $
\State $\hskip2.0em $ return Train $ (k, w, D_k) $
\end{algorithmic}
\end{algorithm}

We based our algorithm on the FedAvg aggregation method by averaging the model weights shared by the clients. We used the warmup model as an initial model starter for the federated learning process. We have detailed the process of our algorithm in the pseudo-code presented in Algorithm \ref{alg:fedavg_data_sharing}. 
 
The collected portions of data samples from users $\zeta$ are used to either train initial model weights $w_0$ or load pre-trained model weights according to the problem's complexity and the available data samples. The transfer learning step is performed by freezing several first layers of the model and keeping the remainder for the model to be trained. Post the training phase, and for the end of $R$ total number of rounds or until model convergence, the server randomly selects a set of available participants $p$. In \textbf{TrainUser} a machine learning training is requested $(k,w_r)$ from each user $k$. The user, in his turn, trains a new model using the personally collected data by his device $D_k$ in federated learning round $w^k_{r+ 1}$. Users' data $D$ are shuffled and split into the mini-batch size of $B$, training and testing sets. The model weights $w$ are updated based on a learning rate $\eta$ specified, while the loss function we used is cross-entropy represented by $\ell$. As for updating the global model weights, we averaged the model weights following the step of the FedAvg algorithm as follows:
  
\begin{equation} \label{1} 
w_{r+1} = \sum ^ k_{ k =1} \frac{n_k}{n} w^k_{r+1} 
\end{equation} 

The weights aggregation creates a superior global model. The model created can differentiate and identify the users' labels participating in the learning process. The aggregated model created is transmitted to the selected users for the following FL round. The continuity of the rounds is determined by the model convergence of a program-specific value.

The cross-entropy loss function $\ell(w;b)$ can be represented as follows:

\begin{equation} \label{} 
\ell(w;b) = \sum_{c=1}^My_{o,c}\log(p_{o,c})
\end{equation}

where $M$ is the number of classes, $y_{o,c}$ is the correct classification of on observation $o$ of a class $c$ and p is the probability predication obtained.

\par \hspace{0pt} \\ \indent
We performed extensive experiments on various datasets, including a facial dataset like the UMDAA-02-FD. The image samples for each user are facial images with a human face in them. Therefore using pre-existing model weights previously trained on identifying human faces can severely boost the accuracy performance. Transfer learning is a helpful approach to transferring the knowledge gained from learning to one of the problems and applying this acquired knowledge to solve another problem. Therefore, we can leverage a previously trained model to boost training performance. In this work, we are using transfer learning as a feature extractor since we are dealing with the facial images of the UMDAA-02-FD dataset. The images represented in these data sets are limited. Thus, we can benefit from the pre-existing open publicly available data set, such as the VGGface2 \cite{cao2018vggface2} dataset that contains millions of images with over 9000+ classes. Therefore, applying transfer learning of the pre-trained model, such as the InceptionResnetV1 \cite{szegedy2017inception} model, over this dataset is a suitable selection since we are dealing with human facial pictures. We use the inception model by training the last five layers and freezing the previous ones in our approach. Therefore, we can customize the model to classify the new classes in the UMDAA-02-FD dataset. The knowledge gained from the pre-trained model weights helps us quickly identify the faces in the images, which improves the model's accuracy instead of training a model from scratch.

\section{Experiments}
\subsection{Implementation and Setup} 
To test the feasibility of our approach, we performed extensive experiments on various publicly available datasets with different characteristics. The datasets that we have used to evaluate our approach are MNIST \cite{deng2012mnist}, CIFAR-10 \cite{krizhevsky2014cifar}, Federated Extended MNIST (FEMNIST) \cite{caldas2018leaf}, and the University of Maryland Active Authentication-02 Face Dataset (UMDAA-02-FD) \cite{Mahbub_Btas2016_UMDAA02}. We processed all datasets to match the case of the continuous authentication format, except UMDAA-02-FD, in which the dataset already fits the continuous authentication use case. Therefore we have created a customized sampling of the data and gave each user a unique set of data that belongs to only one label, matching the realistic case of non-IID data presented in behavioural datasets. The data pre-processing steps are as follows: 
 
\begin{enumerate} 
    \item For all the data available in a dataset, we group all the data and sort them based on the label of the data row.
	\item For every label we have in a dataset, we assign the data that belongs to this label to a client, to which each client has access to only one unique label of the data. The total number of classes for each dataset is represented in table \ref{tab:datasets} alongside the dataset's data feature and total sample size.
	\item The total number of samples is different for every client we have in a dataset. To prepare the client data, we have anticipated a random client distributor, in which we select the clients' numbers and the minimum and the maximum number of samples we need for each client. 
 \end{enumerate}
We had to perform extensive experiments on various datasets with different characteristics to test our hypotheses. We are targeting well-known datasets with numerous data samples and enough clients to perform the clients to match the data/label case we have in continuous authentication. The performance evaluation we made consisted of three different experiments. First, we performed the baseline method of training a ML model in a centralized server, where all the raw sensor data in that experiment are transmitted to a server to train a ML model. We use the term Centralized to represent this experiment. Next, to compare the results of using FL with the FedAvg algorithm, we experimented by applying the federated averaging algorithm and named it FedAvg on the figures. Finally, we applied our approach in a separate experiment using the warmup and transfer knowledge-based federated learning algorithm. We used a transfer learning step in which we used part of the model as a feature extractor network and trained the last layers of the model. We trained this model with the warmup data shared with the server. The initial weights trained from the users' data are much better than starting from random weights compared to the classical federated learning experiments. Therefore, the initially trained model will act as the starter model that the server shares with clients. After the server collects the clients' local models, the weights will be aggregated based on the FedAvg algorithm to create a global model that will work as a classifier for all users. The warmup experiment is explained in more detail in the Framework section. 

Since we are dealing with image data with many dimensions, we had to use a graphics processing unit (GPU) to complete the experiment in a reasonable duration. We used a high-resource Windows platform to process large amounts of data. The machine we performed our experiments on had 16 GB of memory and 16 CPUs at 2.90 GHz speed. We used the CPU for processing the inferences of the experiments and a GPU GTX 1660 Super for processing the training phase of the models. The code we create to run the experiments is inspired by the FedML GitHub repository \cite{chaoyanghe2020fedml}. We utilized the Weights \& Biases (WandB) \cite{wandb} package to visualize the experimental results. We performed several experiments to decide which learning rate was suited for the problem of each dataset and experiment. We used neural network models such as the convolutional neural network (CNN) and the InceptionResnetV1 pre-trained model on the VGGFace2 dataset. The models we used to deal with image classification problems. The models' input is a 2D matrix shape of data resembling the image dimensions in each dataset. In addition, such models have proven to have high accuracy results and model convergence, as proven in the authors' work in the references \cite{8_mcmahan2017communication, szegedy2017inception, chaoyanghe2020fedml}. The InceptionResnetV1 pre-trained model is suitable for the UMDAA-02-FD dataset as it is suitable to detect human faces in images since it is already trained on human faces on the vggface2 dataset \cite{cao2018vggface2}. We have also tested using a transfer learning technique based on feature extraction. The feature extraction boosts the model accuracy with much lower data samples than training the whole model from scratch. Such a technique can help reduce resource consumption and provide more robustness to the model, resulting in fewer communication rounds in the FL training process. We used the InceptionResnetV1 model used in \cite{cao2018vggface2} by fixing the model layers and retraining the last five layers by changing the number of classes the model uses to the number of classes in our datasets. 
We have used the accuracy metric to evaluate the performance of the models over the predicted labels. The accuracy is defined as follows: 
\begin{equation} \label{} 
Accuracy = \frac{TP+TN}{TP+TN+FP+FN}
\end{equation}
Where TP is the true positive, TN is the true negative, FP is the false positive, and FN is the false negative. In addition, we show the averaged accuracy $avg_{acc}$ of the clients $K$ models as follows:

\begin{equation} \label{} 
avg_{acc}=\frac{1}{K} \sum_{k=1}^{k} acc_{k}=\frac{1}{K}\left(acc_{1}+\cdots+acc_{k}\right)
\end{equation} 

In the next subsection, we present our datasets in more detail alongside the experiments and configurations. A notation set for the experiments and graphs is selected. For instance, we use $C$ to represent the total number of participants specified in a FL round. $B$ is the batch size of client data, and $E$ is the total number of local epochs used to train the machine learning model.

\begin{figure}[ht]
\centering
{\includegraphics[width=.45\textwidth]{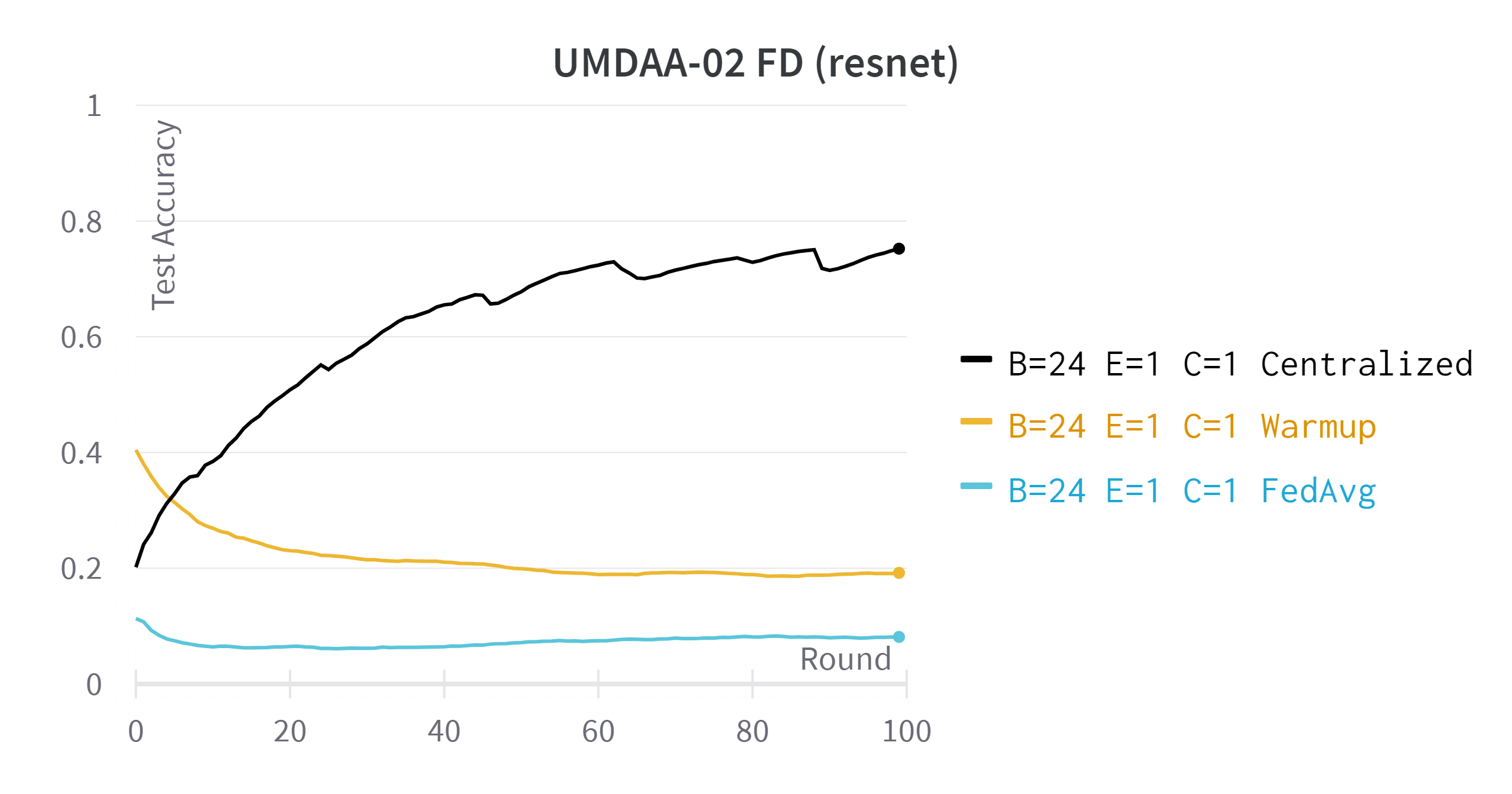}\label{fig:subfig-a}}\hfill
\caption{ Test set accuracy vs. communication rounds
for the UMDAA-02-FD dataset using the resnet56 model.}
\label{fig:umdaa_exp_res}
\end{figure}

\begin{figure}[ht]
\centering
{\includegraphics[width=.45\textwidth]{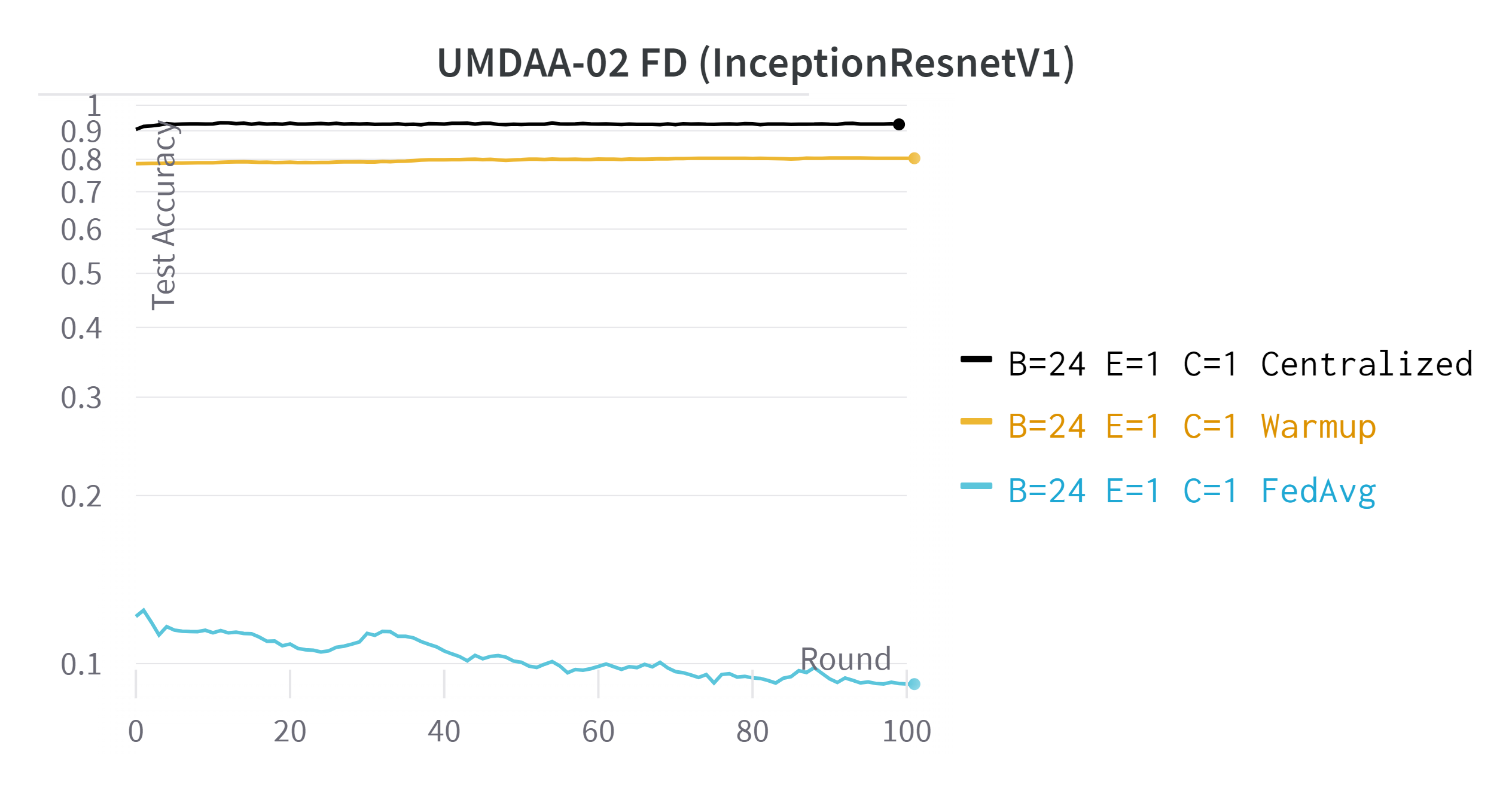}\label{fig:subfig-a}}\hfill
\caption{ Test set accuracy vs. communication rounds
for the UMDAA-02-FD dataset using the InceptionResnetV1 pre-trained model.}
\label{fig:umdaa_exp}
\end{figure}

\par \hspace{0pt} \\ \indent
\textbf{\textit{UMDAA-02-FD}}. The UMDAA-02-FD Face Detection dataset \cite{Mahbub_Btas2016_UMDAA02} consists of images of a very high resolution taken in the wild from partially visible faces taken by the frontal camera of smartphones. The dataset contains 43 users with 33209 images at an interval of 7 seconds. A wide variety of images illuminations, poses, occlusion and facial expressions are captured in these images. In addition, the dataset contains images with no face presented or total black images. The dataset is considered among the hardest datasets for training vision systems due to the considerable variation in the images it contains. The dataset is a perfect match to be considered for a continuous authentication scenario, as the data captured simulate those real case scenarios in a realistic continuous authentication system. In order to process the images promptly, we reduced the dimensions to a fixed width and height, resulting in samples of 128*128 dimension size. We used a sample size of between 400 and 500 images per user for ten clients randomly selected from this dataset to train a ML algorithm and test the dataset for continuous authentication cases. For that purpose, we have experimented with this dataset with strategies of different machine learning models, as follows:
\begin{itemize}
    \item A Resnet ML model trained from scratch.
    \item InceptionResnetV1 pre-trained model on the vggface2 dataset. We performed transfer learning, using the model as a feature extractor, and trained the model's last five layers as a classifier for the network.
\end{itemize}

The resnet machine learning model used is the same as the model implemented by \cite{he2016deep}, as their model show excellent performance in term of model performance and convergence when dealing with complex data input. We tuned the lr as 0.001 for federated and 0.01 for centralized experiments, with B=24 and E=1. We show the details of the configuration and the model used in Figure \ref{fig:umdaa_exp_res}. The plots show that the federated results using the federated averaging algorithm failed to achieve a high-performance result. However, the warmup approach applied using the resnet model increased accuracy by only 10\% after 100 rounds. Moreover, the centralized experiment showed an accuracy below 80\% after 100 rounds.
In comparison, we used a transfer learning technique of the model InceptionResnetV1 as a feature extractor by taking advantage of its pre-trained weights to identify the faces in images of the UMDAA-02-FD dataset. We set the number of classes to match the number of classes in our dataset. We used a low learning rate of 0.001 for both the federated and centralized experiments. The batch is set to be 24, and epochs are equal to 1 in each round. By referring to the graph in Figure \ref{fig:umdaa_exp}, it can be noted that the FedAvg has failed to classify the data samples as the model accuracy drops below 10\%, in contrast to the warmup experiment with 80\% where we can see it can closely match the centralized experiment. A considerable performance increase of 70\% using only 5\% of the data for the warmup model. Therefore using the pre-trained model has severely increased the model's performance using our warmup approach, with a difference of 60\% increase from the resnet56 model.  

\par
\begin{figure}[htp]

\centering
{\includegraphics[width=.45\textwidth]{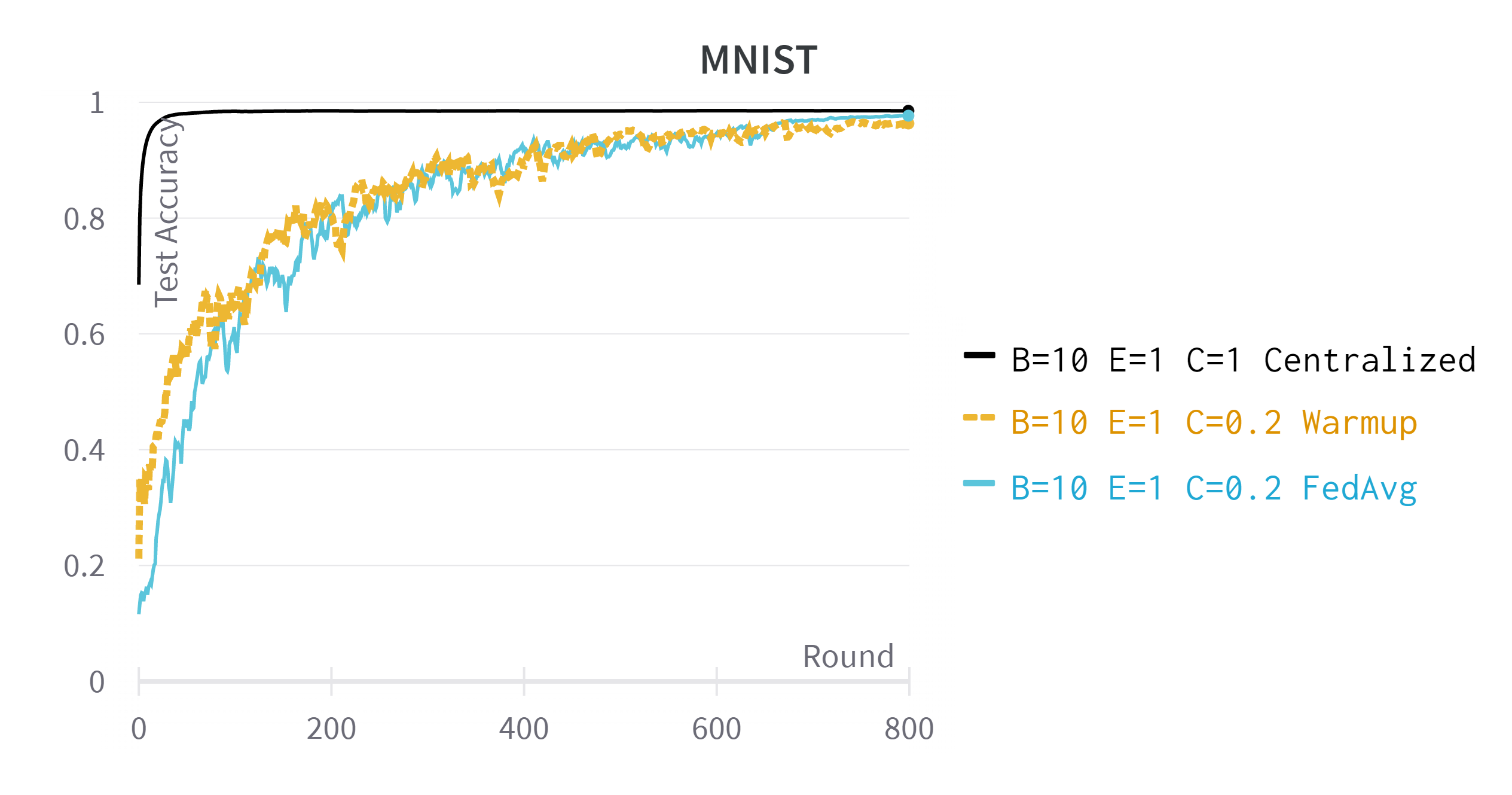}\label{fig:subfig-a}}\hfill
\caption{ Test set accuracy vs. communication rounds
for the MNIST dataset. Each client has a sample size range of [800-1000].}
\label{fig:mnist_exp}
\end{figure}

 \hspace{0pt} \\ \indent
\textbf{\textit{MNIST}}. The MNIST dataset \cite{deng2012mnist} is one of the numerous used datasets for training visual ML systems, as it contains 784 features for each data sample with over 70,0000 data sample size. We have used this dataset to create a custom data distribution by considering each label as an independent client. We used a random samples distributor to create partitions of 10 clients with labels from zero to nine, each with a random number of samples between 800 and 1000 data images. To train this dataset, we have a CNN ml model, similar to the one used in \cite{8_mcmahan2017communication}, in which CNNs are very popular in training for visual systems since they deal with a high number of dimensions as input, which is the case in the data samples of MNIST. Referring to the graphs in Figure \ref{fig:mnist_exp}, we see that for the MNIST dataset, a small data distribution \ref{fig:mnist_exp} per user is sufficient to identify the user accurately.
The warmup diagram refers to the exchange strategy where we use only 5\% of the data from each client to create an initialized model. In addition, we performed experiments for 800 communication rounds to achieve model convergence. In each round of federated learning, we select only 20\% of the number of random clients for MNIST. We assume that only a few clients are ready for a particular round of federated learning. The results show that when using the FedAvg algorithm, we achieved approximately 97\% accuracy for the distribution of single-labelled data across ten clients using the MNIST dataset. We also see that the warmup model slightly improves model convergence and achieves similar results to the FedAvg algorithm. We use different configurations to run the MNIST experiments with epochs set to 1, with a B=10 and a learning rate (lr) equal to 0.1 and 0.001 for federated and centralized experiments, respectively. Based on the MNIST experiment in Figure \ref{fig:mnist_exp}, we see a  minor difference in the final round between FedAvg and centralized experiments. The model with the federated results achieves a convergence with 96.22\% accuracy compared to the centralized result with 98.53\%.

\begin{figure}[htp]
  \centering
  \includegraphics[width=.45\textwidth]{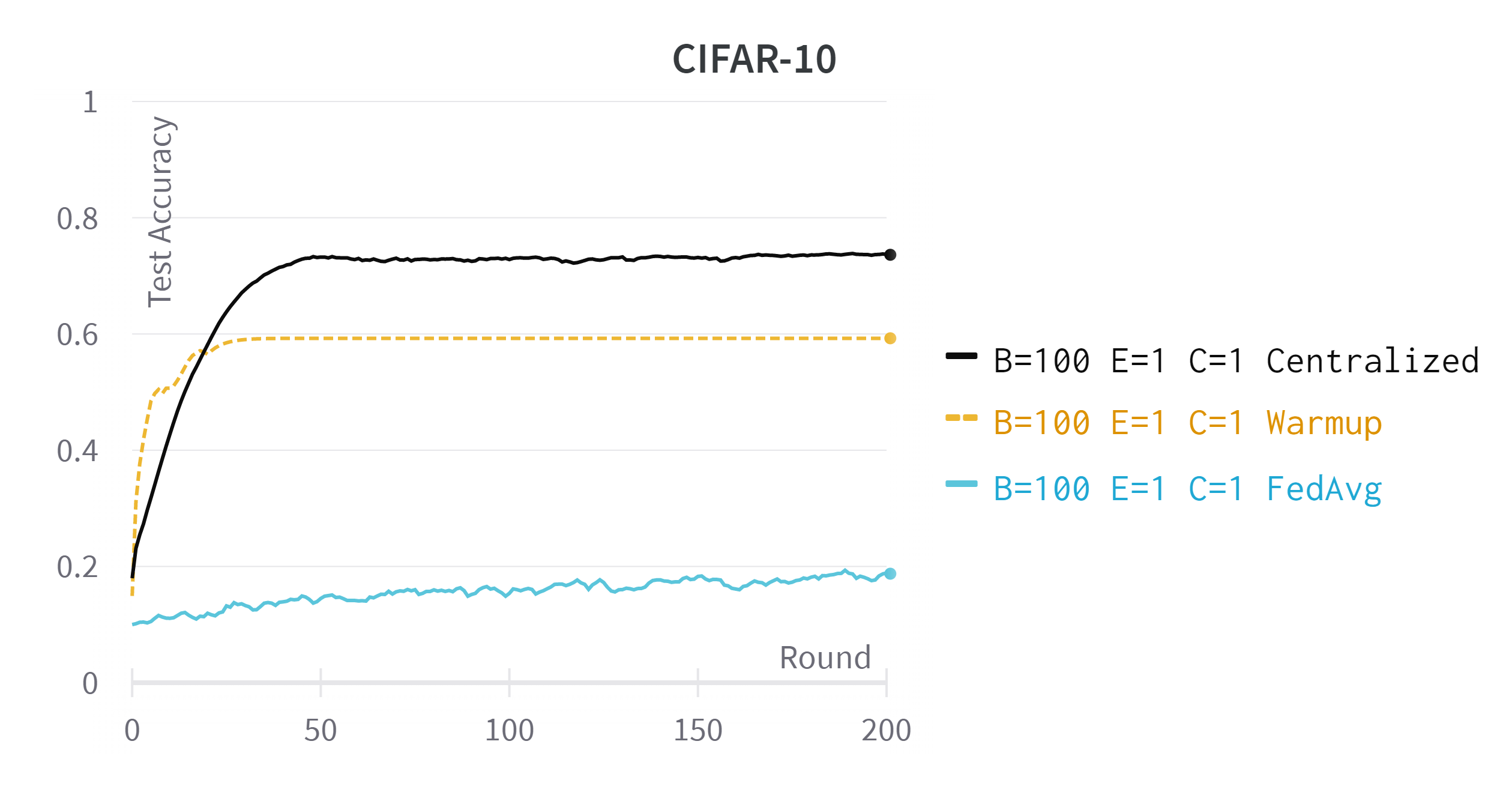}\hfill
    \caption{Accuracy test results for the Cifar-10 dataset.}
    \label{fig:cifar-10}
\end{figure}

\par \hspace{0pt} \\ \indent
\textbf{\textit{CIFAR-10}}. The CIFAR-10 \cite{krizhevsky2014cifar} dataset is also a widely known dataset that we have used in our experiments to validate our approach. The dataset contains images with 32*32 pixels per data sample. It has ten classes ranging from animals to machines. The dataset is also commonly used in training computer vision algorithms. We have used our data distributor to create partitions of ten different clients, with 6000 images per client data sample. We perform the CIFAR-10 experiments with a 6 conv2D layers model for both experiments. We tried working with a CNN with fewer layers, but the results were inferior. We set E=1 and B=10. We performed several experiments and chose an lr equal to 0.1 and 0.001 for the federated and centralized experiments, respectively. We show the different experiments result in Figure \ref{fig:cifar-10}. We plot the results of the accuracy test experiments with the Cifar-10 dataset. We showed that the centralized experiment achieves excellent results with just a few rounds compared to warmup and FedAvg experiments. However, the warmup experiment demonstrated that with only 5\% of the data shared with a server, we achieved a 30\% increase in accuracy performance.

\par
\begin{figure}[htp]
  \centering
  \includegraphics[width=.45\textwidth]{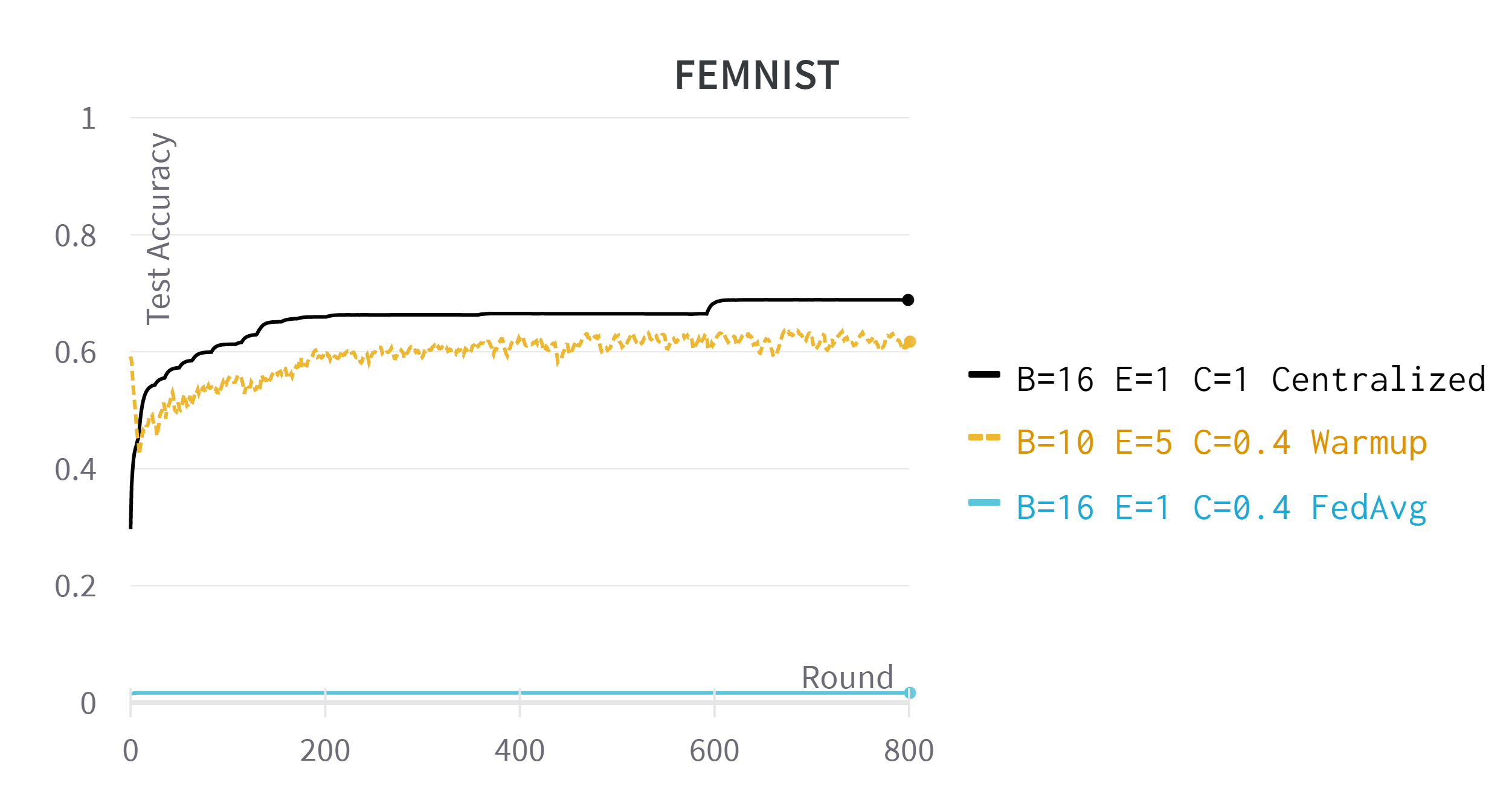}\hfill
    \caption{Accuracy test scores for the FEMNIST dataset.}
    \label{fig:F-emnist}
\end{figure}

 \hspace{0pt} \\ \indent
\textbf{\textit{Federated Extended MNIST (FEMNIST)}}. The FEMNIST dataset is an extension of the MNIST \cite{caldas2018leaf}. It contains over 700,000 data samples. Each sample is an image of 28*28 pixels. The images represent digits from zero to nine, in addition to English alphabetic letters with lower and capitalized cases. The total number of classes presented in this dataset is 62 labels. We created a custom partition for our experiment. We group the data samples of each class separately, and we distribute a random number of samples in the range of [2000-3000] for each client, each with a unique label. Since we are dealing with image datasets, we used the CNN machine learning model, which showed promising results when training with visual datasets. The learning rate we chose for the federated experiments is fixed at 0.01 and 0.001 for the centralized experiment. We used only 5\% of the data samples per user to train an initial model for the warmup experiment. From the plots in Figure \ref{fig:F-emnist} we see that the warmup model created by using only a 5\% portion of the data from the user has provided an initial boost of 60\% accuracy compared to the classical federated experiment with FedAvg algorithm, where the accuracy of the model remains under 2\%.

\begin{table}[]
\centering
\begin{tabular}{|c|c|c|c|c|}
\hline
\textbf{Dataset}                                            & \textbf{\begin{tabular}[c]{@{}c@{}}\#Train \\ Examples\end{tabular}} & \textbf{\begin{tabular}[c]{@{}c@{}}\#Test \\ Examples\end{tabular}} & \textbf{\#Features} & \textbf{\#Classes} \\ \hline
UMDAA-02-FD                                                    & 26,566                                                               & 6,641                                                              & 16,384                & 43                 \\ \hline
CIFAR-10                                                    & 50,000                                                               & 10,000                                                              & 1,024                & 10                 \\ \hline
\begin{tabular}[c]{@{}c@{}}FEMNIST\end{tabular} & 671,585                                                               & 77,483                                                               & 784                 & 62                 \\ \hline
MNIST                                                       & 60,000                                                                & 10,000                                                               & 784                 & 10  
\\ \hline
\end{tabular}
\caption{Datasets details}
\label{tab:datasets}
\end{table}

\subsection{Discussion}

The preexisting works focused on devising a central server and transferring all clients' data to create a centralized machine learning model that can authenticate the user devices, exposing private data to malicious attacks. We affirm that applying a FL approach to behavioural data can address the privacy and security of the data. We also show that the nature of individual behavioural data collected from personal smartphones and IoT devices has a unique shape, and they are in the form of non-IID. Therefore, each client has a unique label for a group of samples, and non of the users share data. The uniqueness of data labels between the clients imposes a severe problem for federated learning when applying the FedAvg algorithm. The weight divergence of the ML model increases between the models, leading to a decreased global model created when averaging on the server. However, we showed that a federated learning approach with an initial backbone model could help reduce the divergence of the model weights, causing a model performance increase. Furthermore, the transfer learning approach showed superior results compared to a classifier trained from scratch. Therefore, we take advantage of federated learning characteristics by having one global model that preserves the user's private data while maintaining high model accuracy. Extensive experiments showed that our novel strategy could perform well using the FedAvg algorithm, with an additional 40\% increase on the Cifar-10, 60\% on FEMNIST and 70\% on UMDAA-02-FD datasets.
In contrast to the work of \cite{14_zhao2018federated}, our implementation did not include shared data between the clients. We could satisfy a high model performance without compromising users' private data. Extensive experiments showed that our models could achieve high accuracy results, and our feature extractor using a pre-trained model showed that it can further increase the accuracy results. 

\section{Conclusion and future work}
Continuous authentication adds an essential line of defence for private user data, making it hard for adversaries to compromise the authentication of a system. In this work, we addressed the issue of transferring full user private data to an external system by proposing a novel privacy-preserving federated learning approach that can achieve the desired results while maintaining the privacy and security of users' data. Moreover, we addressed the problem of implementing a federated learning approach using non-IID data presented in individual unique behavioural data. We also show that FedAvg fails to achieve high accuracy when exposed to such behavioural data, leading to a severe reduction in model accuracy. Our warmup transfer-knowledge strategy can increase the model's accuracy while maintaining minimal shared data. In addition, we demonstrated that the transfer learning technique based on the feature extractor method offered an advantage when dealing with facial images. The experimental results showed model accuracies increase by 40\% for CIFAR-10, 60\% for FEMNIST and 70\% for UMDAA-02-FD. In future work, we aim to examine personalized models for federated learning to gain better accuracy results for each user model end while maintaining fewer communication rounds to decrease the cost of communication for the clients.

\section*{Acknowledgement}
This work was cofounded by IPTOKI (www.iptoki.com) and MITACS (www.mitacs.ca). 

\bibliographystyle{IEEEtran}
\bibliography{references}
 
\begin{IEEEbiography}
[{\includegraphics[width=1in,height=1.25in,clip,keepaspectratio]{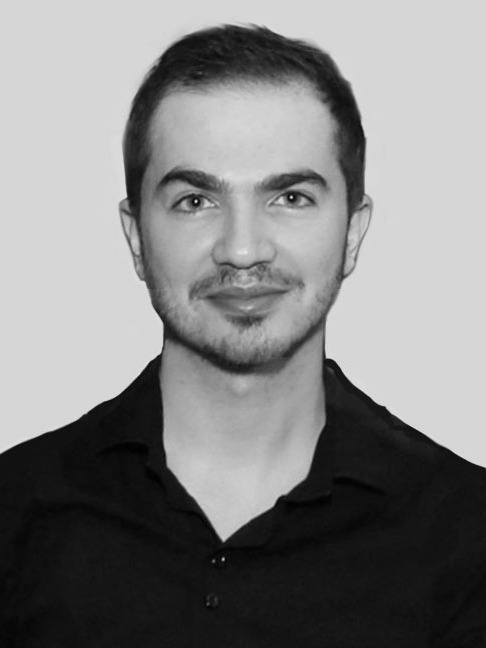}}]
{\textbf{\textit{\textbf{Mohamad Wazzeh}}}}
Received the M.S. degree in business computing from the Lebanese University in Tripoli, Lebanon, in 2016. He is currently pursuing a Ph.D. degree with the École de Technologie Supérieure, Montreal, QC, Canada. He is a data scientist intern with IPtoki, Montreal. His research interests include continuous authentication, distributed learning, machine learning, and security. He is a reviewer in several prestigious conferences and journals.
\end{IEEEbiography}
\vskip 0pt plus -1fil
\begin{IEEEbiography}
[{\includegraphics[width=1in,height=1.25in,clip,keepaspectratio]{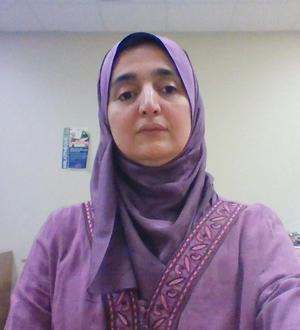}}]
{\textbf{\textit{\textbf{Hakima Ould-Slimane}}}}
Obtained her Ph.D. degree in Computer Science from Laval University, Quebec, Canada. She is currently a professor at the department of mathematics and computer science at Universite de Quebec a Trois-Rivieres (UQTR, Trois-Rivieres, Canada). Her research interests include mainly: information security, cyber resilience, homomorphic encryption, federated learning, preserving data privacy in smart environments, machine learning based intrusion detection, access control, optimization of security mechanisms and security of social networks.
\end{IEEEbiography}
\vskip 0pt plus -1fil
\begin{IEEEbiography}
[{\includegraphics[width=1in,height=1.25in,clip,keepaspectratio]{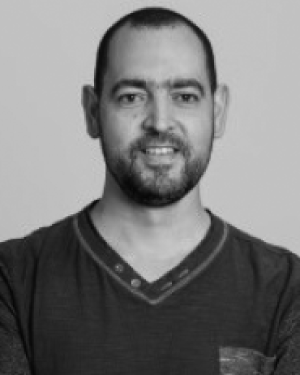}}]
{\textbf{\textit{\textbf{Chamseddine Talhi}}}}
Received the Ph.D. degree in computer science from Laval University, Quebec, QC, Canada, in 2007. He is an Associate Professor with the Department of Software Engineering and IT, ÉTS, University of Quebec, Montreal, QC, Canada. He is leading a research group that investigates smartphone, embedded systems, and IoT security. His research interests include cloud security and secure sharing of embedded systems.
\end{IEEEbiography}
\vskip 0pt plus -1fil
\begin{IEEEbiography}[{\includegraphics[width=1in,height=1.25in,clip,keepaspectratio]{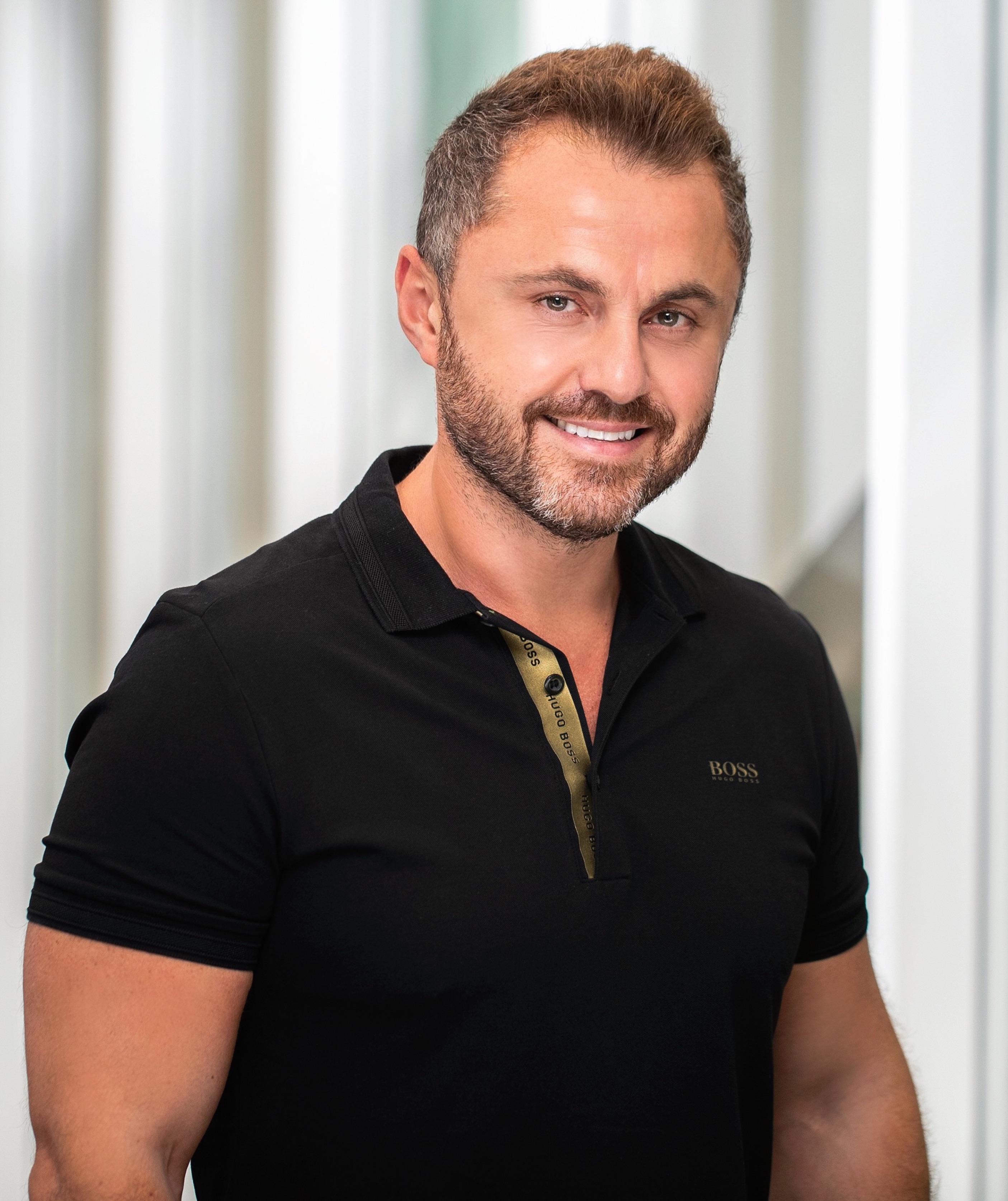}}]%
{\textbf{\textit{\textbf{Azzam Mourad}}}}received his M.Sc. in CS from Laval University, Canada (2003) and Ph.D. in ECE from Concordia University, Canada (2008). He is currently Professor of Computer Science and Founding Director of the Cyber Security Systems and Applied AI Research Center with the Lebanese American University, Visiting Professor of Computer Science with New York University Abu Dhabi and Affiliate Professor with the Software Engineering and IT Department, Ecole de Technologie Superieure (ETS), Montreal, Canada. His research interests include Cyber Security, Federated Machine Learning, Network and Service Optimization and Management targeting IoT and IoV, Cloud/Fog/Edge Computing, and Vehicular and Mobile Networks. He has served/serves as an associate editor for IEEE Transactions on Services Computing, IEEE Transactions on Network and Service Management, IEEE Network, IEEE Open Journal of the Communications Society, IET Quantum Communication, and IEEE Communications Letters, the General Chair of IWCMC2020, the General Co-Chair of WiMob2016, and the Track Chair, a TPC member, and a reviewer for several prestigious journals and conferences. He is an IEEE senior member.
\end{IEEEbiography}
\vskip 0pt plus -1fil
\begin{IEEEbiography}
[{\includegraphics[width=1in,height=1.25in,clip,keepaspectratio]{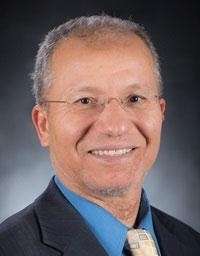}}]%
{\textbf{\textit{\textbf{Mohsen Guizani}}}}
(Fellow, IEEE) received the BS (with distinction), MS and PhD degrees in Electrical and Computer engineering from Syracuse University, Syracuse, NY, USA in 1985, 1987 and 1990, respectively. He is currently a Professor of Machine Learning and the Associate Provost at Mohamed Bin Zayed University of Artificial Intelligence (MBZUAI), Abu Dhabi, UAE. Previously, he worked in different institutions in the USA. His research interests include applied machine learning and artificial intelligence, Internet of Things (IoT), intelligent autonomous systems, smart city, and cybersecurity. He was elevated to the IEEE Fellow in 2009 and was listed as a Clarivate Analytics Highly Cited Researcher in Computer Science in 2019, 2020 and 2021. Dr. Guizani has won several research awards including the “2015 IEEE Communications Society Best Survey Paper Award”, the Best ComSoc Journal Paper Award in 2021 as well five Best Paper Awards from ICC and Globecom Conferences. He is the author of ten books and more than 800 publications. He is also the recipient of the 2017 IEEE Communications Society Wireless Technical Committee (WTC) Recognition Award, the 2018 AdHoc Technical Committee Recognition Award, and the 2019 IEEE Communications and Information Security Technical Recognition (CISTC) Award. He served as the Editor-in-Chief of IEEE Network and is currently serving on the Editorial Boards of many IEEE Transactions and Magazines. He was the Chair of the IEEE Communications Society Wireless Technical Committee and the Chair of the TAOS Technical Committee. He served as the IEEE Computer Society Distinguished Speaker and is currently the IEEE ComSoc Distinguished Lecturer. 
\end{IEEEbiography}

\end{document}